\documentclass[10pt,twocolumn,letterpaper]{article}

\usepackage[article]{cvpr}

\usepackage{graphicx}
\usepackage{amsmath}
\usepackage{amssymb}
\usepackage{booktabs}
\usepackage{multirow}
\usepackage{array}
\usepackage{makecell}
\usepackage{subcaption}

\definecolor{cvprblue}{rgb}{0.21,0.49,0.74}
\usepackage[pagebackref,breaklinks,colorlinks,citecolor=cvprblue]{hyperref}

\graphicspath{{img/}{figures/}{../img/}{./}{../}}

\def\paperID{}
\def\confName{CVPR}
\def\confYear{}

\begin{document}

\title{From Code to Prediction: Fine-Tuning LLMs for Neural Network\\Performance Classification in NNGPT}

\author{
     Mahmoud Hanouneh,
    \hspace{0.2cm} Radu Timofte,
    \hspace{0.2cm} Dmitry Ignatov\\
    \small{Computer Vision Lab, CAIDAS \& IFI, University of W\"urzburg, Germany}
}
\maketitle

\begin{abstract}
Automated Machine Learning (AutoML) frameworks increasingly leverage Large Language Models (LLMs) for tasks such as hyperparameter optimization and neural architecture code generation. However, current LLM-based approaches focus on generative outputs and evaluate them by training the produced artifacts. Whether LLMs can learn to \emph{reason} about neural network performance across datasets remains underexplored. We present a classification task integrated into the NNGPT framework, in which a fine-tuned LLM predicts which of two image classification datasets a given neural network architecture achieves higher accuracy on. The task is built on the LEMUR dataset, which provides standardized PyTorch implementations with reproducible performance metrics. Three prompt configurations of increasing difficulty are evaluated: a normalized-accuracy baseline (trivially reaching 100\%), a metadata-enriched prompt replacing accuracies with dataset properties, and a code-only prompt presenting only architecture source code and dataset names. Using DeepSeek-Coder-7B-Instruct fine-tuned with LoRA, the code-only prompt reaches 80\% peak accuracy over 15 epochs, while the metadata prompt peaks at 70\%. Per-dataset analysis reveals complementary strengths: metadata excels for datasets with distinctive properties (CelebA-Gender at 90.9\%) but degrades for overlapping characteristics, whereas the code-only prompt shows more balanced performance. A comparison with DeepSeek-Coder-1.3B confirms that model capacity affects this form of architectural reasoning. The results establish that LLMs can be fine-tuned to predict cross-dataset suitability from neural network code, suggesting that architecture source code contains richer discriminative signal than dataset metadata alone.
\end{abstract}


\section{Introduction}
\label{sec:intro}

The rapid progress of deep learning has produced increasingly powerful neural network architectures for computer vision; however, designing, tuning, and benchmarking these models remains a slow, resource-intensive process that depends heavily on expert knowledge~\cite{ABrain.NN-Dataset}. The choice of architecture and hyperparameters directly determines model performance, yet traditional approaches rely on manual experimentation or computationally expensive search procedures~\cite{ABrain.HPGPT}. Automated Machine Learning (AutoML) has emerged to address these challenges by automating the design and optimization of ML pipelines in a data-driven way, aiming to free experts for higher-level tasks and enable novice users to train well-performing models~\cite{Gijsbers2024AMLB}. Classical AutoML techniques---grid search, random search, Bayesian optimization, and evolutionary strategies---explore large configuration spaces but require numerous training iterations, leading to massive computational overhead for complex deep learning models~\cite{ABrain.HPGPT}.

Large Language Models (LLMs) have recently demonstrated the ability to capture complex patterns across a variety of tasks and, with fine-tuning, can acquire new capabilities to solve previously unseen problems~\cite{Liu2025AgentHPO,ABrain.NN-Dataset}. Their application to AutoML delivers a paradigm shift: instead of iteratively searching for optimal configurations, LLMs can learn from historical experimental data and generate solutions in a single forward pass, significantly reducing computational cost~\cite{ABrain.NNGPT}. The NNGPT framework exemplifies this approach by integrating LLMs into the machine learning pipeline to generate or optimize neural network architectures, leveraging learned representations of past experiments from the LEMUR Neural Network Dataset~\cite{ABrain.NN-Dataset}. Unlike traditional datasets that focus solely on input data, LEMUR treats the neural network code itself as data, providing a unified collection of PyTorch implementations across multiple tasks with consistent evaluation pipelines and structured logging of results~\cite{ABrain.NN-Dataset}. This structured representation enables LLMs to learn relationships between architectures, hyperparameters, and performance outcomes.

Despite these advances, a gap remains in how LLMs are evaluated within AutoML frameworks. Existing LLM-based approaches focus on generative tasks---producing neural network code or predicting numerical hyperparameter values---and their evaluation relies on training the generated outputs and measuring downstream accuracy~\cite{ABrain.NNGPT,ABrain.HPGPT}. This paradigm does not address a fundamentally different question: whether an LLM can learn to \emph{reason} about neural network performance across datasets, classifying which of two datasets a given architecture performs better on without generating or training any model. Prior to this work, the NNGPT pipeline supported only code-generation output types and had no mechanism for classification tasks where the LLM's output is a discrete label rather than executable code. Similarly, benchmarking studies such as AMLB~\cite{Gijsbers2024AMLB} compare AutoML frameworks across tasks and data characteristics but do not test an LLM's capacity for meta-level reasoning about model--dataset compatibility, and agent-based approaches such as AgentHPO~\cite{Liu2025AgentHPO} focus on iteratively refining hyperparameters rather than predicting cross-dataset performance differences. This leaves open the question of whether LLMs, when fine-tuned on structured performance data from repositories such as LEMUR, can develop a meaningful understanding of the relationship between neural network architectures and dataset properties.

This work addresses this gap by presenting the design, implementation, and evaluation of a classification task integrated into the NNGPT framework. Specifically, given a neural network's source code and two candidate dataset names, the fine-tuned LLM must predict which dataset the network achieves higher accuracy on. The contributions of this work are threefold: (1)~a novel classification task formulation within the NNGPT framework that extends LLM-based AutoML beyond generative outputs to reasoning about architecture--dataset relationships; (2)~a structured ablation study across three prompt variants that progressively remove explicit performance signals, revealing that architecture source code carries richer discriminative signal than dataset metadata; and (3)~empirical evidence that fine-tuned LLMs can predict cross-dataset suitability from source code alone, achieving 80\% classification accuracy with DeepSeek-Coder-7B \cite{deepseekai2024deepseekv3technicalreport}.

\section{Related Work}
\label{sec:related}

\subsection{LLM-Based Hyperparameter Optimization}
\label{sec:related_hpo}

Recent work has explored LLMs as alternatives to traditional hyperparameter optimization methods. Kochnev et al.~\cite{ABrain.HPGPT} demonstrated that fine-tuned LLMs can match or outperform Optuna on hyperparameter tuning tasks while requiring fewer evaluations. AgentHPO~\cite{Liu2025AgentHPO} introduces a multi-agent framework in which a Creator agent generates hyperparameter configurations and an Executor agent trains and evaluates them iteratively, achieving competitive results across standard benchmarks. These approaches optimize \emph{parameters} but do not address whether LLMs can predict cross-dataset performance differences.

\subsection{LLM-Driven Neural Architecture Generation}
\label{sec:related_nas}

NNGPT~\cite{ABrain.NNGPT} represents a paradigm shift in neural architecture search by fine-tuning LLMs on structured architecture--performance data from the LEMUR dataset to generate executable PyTorch code in a single forward pass. The framework employs a closed-loop LangGraph-based agent orchestration where generated architectures are trained, evaluated, and fed back into LoRA fine-tuning. Khalid et al.~\cite{ABrain.Architect} extend this direction by incorporating novelty detection to encourage architecturally diverse generations. While these systems generate architectures, they do not classify or predict performance across datasets---the task addressed in this work.

\subsection{Benchmarking and Datasets for AutoML}
\label{sec:related_bench}

Standardized benchmarking is essential for reproducible AutoML research. The AMLB benchmark~\cite{Gijsbers2024AMLB} provides a unified evaluation framework for comparing AutoML systems, while AutoMLBench~\cite{Eldeeb2024AutoMLBench} systematically evaluates the impact of design decisions such as search space size and meta-learning. The LEMUR dataset~\cite{ABrain.NN-Dataset, ABrain.LEMUR2} addresses a complementary need by treating neural network \emph{code} as data: it provides a unified collection of PyTorch implementations across multiple tasks with consistent evaluation pipelines and structured logging of experimental results. This structured representation is particularly important for LLM-based approaches, as it allows models to learn relationships between architectures, hyperparameters, and performance outcomes. LEMUR provides the data foundation that makes the classification task presented in this work possible.

\section{Methodology}
\label{sec:methodology}

\subsection{Task Formulation}
\label{sec:task}

The classification task is formulated as a constrained text generation problem. Given a neural network architecture (represented as truncated PyTorch source code) and the names of two image classification datasets, the fine-tuned LLM must generate a single token: the name of the dataset on which the architecture achieves higher accuracy. The model receives a structured prompt as plain text and outputs a dataset name as a free-form string, evaluated against the ground truth via cascaded string matching (exact, substring, and normalized substring).

The ground-truth label for each pair is derived from normalized accuracy values computed over the LEMUR corpus. For a dataset $d$, the normalized accuracy of model $m$ is defined as:
\begin{equation}
\text{norm\_acc}(m,d) = \frac{\mathrm{acc}(m,d,e)}{\max_{m'} \mathrm{acc}(m', d, e)}
\label{eq:norm_acc}
\end{equation}
where the maximum is taken over all models evaluated at the fixed reference epoch $e{=}5$. For a pair $(d_1, d_2)$, the ground-truth label is the dataset with the higher normalized accuracy. This normalization ensures that cross-dataset comparisons reflect relative model suitability rather than absolute dataset difficulty~\cite{ABrain.NN-Dataset,Gijsbers2024AMLB}.

The pairwise formulation yields $\binom{n}{2}$ distinct dataset pairs per architecture, providing substantially more training signal than a single multi-class label. Training samples are constructed via SQL self-joins with an ordering constraint ($d_1.\text{id} < d_2.\text{id}$) that eliminates duplicate and reflexive pairs~\cite{ABrain.NNGPT}.

\subsection{Prompt Design}
\label{sec:prompts}

Three prompt configurations form a structured ablation that progressively removes explicit performance signals:

\noindent\textbf{V1 -- Normalized-accuracy baseline} (N=10): provides the neural network source code, both dataset names, and both normalized accuracy values. This serves as a pipeline validation since the LLM need only compare two numbers.

\noindent\textbf{V2 -- Metadata-enriched} (N=30): replaces accuracy values with dataset properties (number of training images, image size, number of channels, number of classes). The LLM must infer relative performance from the interplay between architectural features and dataset properties.

\noindent\textbf{V3 -- Code-only} (N=30): presents only the neural network source code and dataset names, with no accuracy or metadata information. The model must predict the better-suited dataset based solely on its understanding of the architecture's structure.

All configurations share: \texttt{output\_type=classification}, 20-token generation limit, and 2,000-character code truncation. The ground-truth label \texttt{better\_dataset} is excluded from input fields and injected only on the output side of training pairs to prevent data leakage.

\subsection{Pipeline Integration}
\label{sec:pipeline}

The classification task reuses NNGPT's LangGraph-based multi-agent orchestration (Fine-tuner $\rightarrow$ Generator $\rightarrow$ Evaluator)~\cite{ABrain.NNGPT}, but routes outputs through a dedicated classification evaluator rather than the neural network training branch. This preserves the closed-loop self-improvement property where execution logs feed back into LoRA fine-tuning. The same LoRA configuration (rank=32, $\alpha$=32, dropout=0.05), cosine learning rate scheduler, and three inner training epochs apply to classification as to code generation, ensuring consistency. For the DeepSeek-1.3B experiments, ONNX Runtime~\cite{onnx} acceleration is used with a separate tuning script.

\section{Experiments}
\label{sec:experiments}

\subsection{Experimental Setup}
\label{sec:protocol}

All experiments were conducted on the Computer Vision Lab Kubernetes cluster at the University of W\"urzburg using NVIDIA GeForce RTX~4090 GPUs (24\,GB VRAM), via the AI~Linux docker image\footnote{AI Linux: \scriptsize\url{https://hub.docker.com/r/abrainone/ai-linux}}. The primary LLM was DeepSeek-Coder-7B-Instruct-v1.5, fine-tuned with LoRA (rank=32, $\alpha$=32, dropout=0.05) using a cosine learning rate scheduler and three inner training epochs per outer fine-tuning cycle~\cite{ABrain.HPGPT}. An additional experiment used DeepSeek-Coder-1.3B-Instruct with ONNX Runtime acceleration for the code-only variant. Classification accuracy was computed at each epoch as the proportion of correctly predicted dataset names out of total test samples.

\subsection{Results}
\label{sec:results}

Table~\ref{tab:summary} summarizes peak accuracy across all configurations. The normalized-accuracy baseline saturates at 100\%, confirming that the task is trivially solvable when explicit performance values are provided. The code-only prompt achieves the highest non-trivial accuracy at 80\%.

\begin{table}[t]
    \centering
    \fontsize{7.5}{9}\selectfont
    \begin{tabular}{l c c c c c}
        \toprule
        \textbf{Prompt} & \textbf{Model} & \textbf{N} & \textbf{Peak} & \textbf{Ep.} & \textbf{Run} \\
        \midrule
        Norm.\ Acc.  & 7B   & 10 & 100.0\% & 9      & 12 \\
        Code-Only    & 7B   & 30 & 80.0\%  & 15     & 16 \\
        Metadata     & 7B   & 30 & 70.0\%  & 8, 13  & 15 \\
        Code-Only    & 1.3B & 30 & 70.0\%  & 22, 23 & 29 \\
        \bottomrule
    \end{tabular}
    \caption{Peak classification accuracy across prompt variants and model sizes. Peak = peak accuracy; Ep.\ = epoch of peak; Run = total epochs.}
    \label{tab:summary}
\end{table}

\subsubsection{Code-Only Prompt.}
With only neural network source code and dataset names, DeepSeek-7B started at 20\% accuracy at epoch~0---notably below the 50\% random baseline, indicating a systematic bias in the pre-trained model's prior probability over dataset name tokens. Despite this unfavorable starting point, the model showed a clear learning trajectory, reaching 80\% at epoch~15 (24/30 correct). Table~\ref{tab:code_only_epochs} shows the epoch-by-epoch progression.

\begin{table}[t]
    \centering
    \fontsize{7.5}{9}\selectfont
    \begin{tabular}{cccc}
        \toprule
        \textbf{Ep.} & \textbf{Corr.} & \textbf{Tot.} & \textbf{Acc.} \\
        \midrule
        0 & 6 & 30 & 20.0\% \\
        2 & 14 & 30 & 46.7\% \\
        4 & 15 & 30 & 50.0\% \\
        6 & 17 & 30 & 56.7\% \\
        9 & 20 & 30 & 66.7\% \\
        10 & 21 & 30 & 70.0\% \\
        12 & 23 & 30 & 76.7\% \\
        13 & 23 & 30 & 76.7\% \\
        15 & 24 & 30 & 80.0\% \\
        \bottomrule
    \end{tabular}
    \caption{Selected epochs for the code-only prompt (DeepSeek-7B, N=30). Accuracy rises from 20\% to 80\% over 15 fine-tuning epochs.}
    \label{tab:code_only_epochs}
\end{table}

The same configuration evaluated with DeepSeek-1.3B using ONNX Runtime showed substantially more volatile behavior, oscillating between 23\% and 70\% and peaking at 70\% at epochs~22--23. Unlike the 7B model's steady upward trajectory, the 1.3B model showed no continuous improvement trend in its first 13 epochs.

\begin{figure}[t]
    \centering
    \includegraphics[width=\columnwidth]{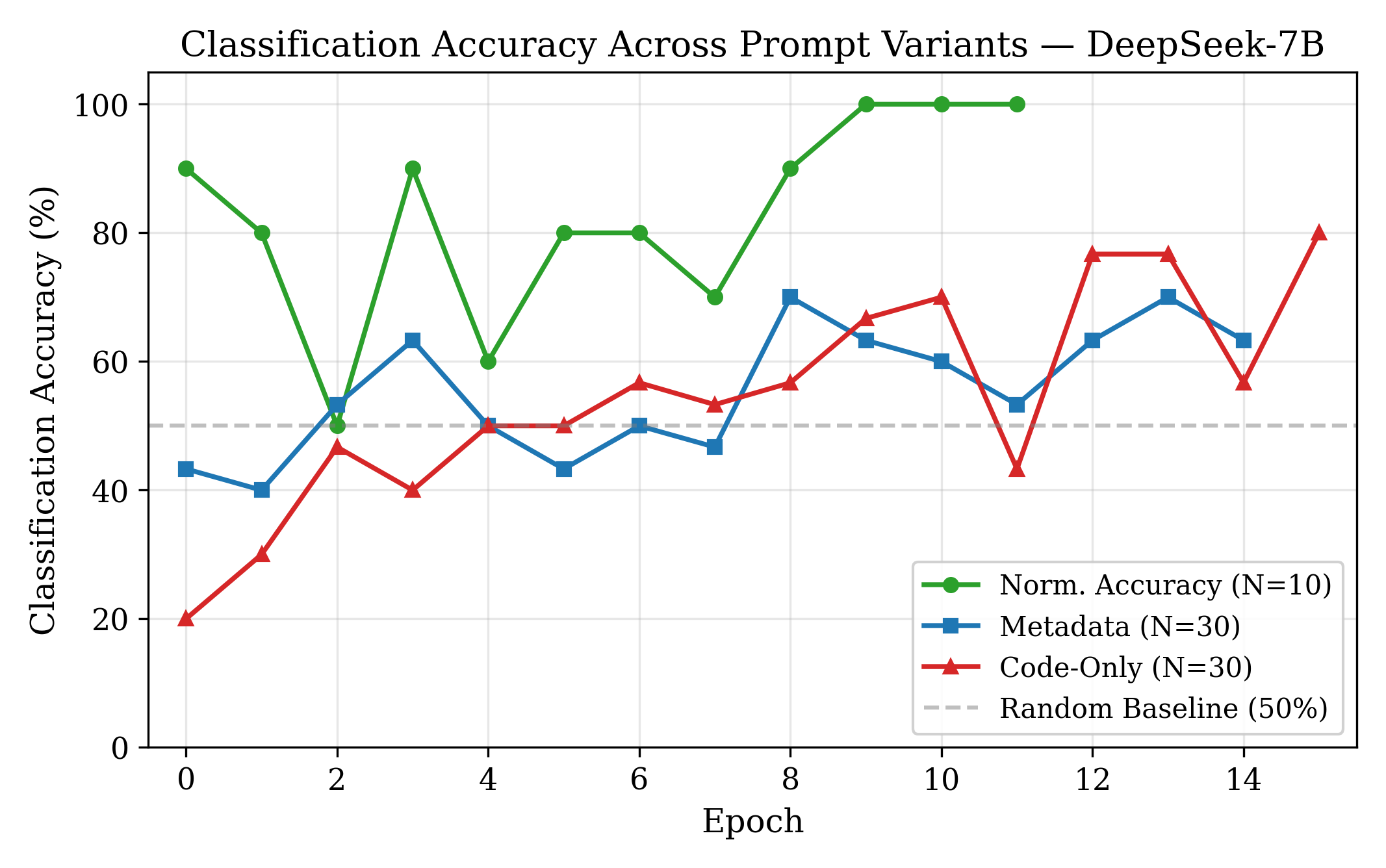}
    \caption{Classification accuracy over epochs for all three prompt variants (DeepSeek-7B). The normalized-accuracy baseline saturates at 100\%, while code-only and metadata prompts show gradual learning above the 50\% random baseline.}
    \label{fig:accuracy_variants}
\end{figure}

\begin{figure}[t]
    \centering
    \includegraphics[width=\columnwidth]{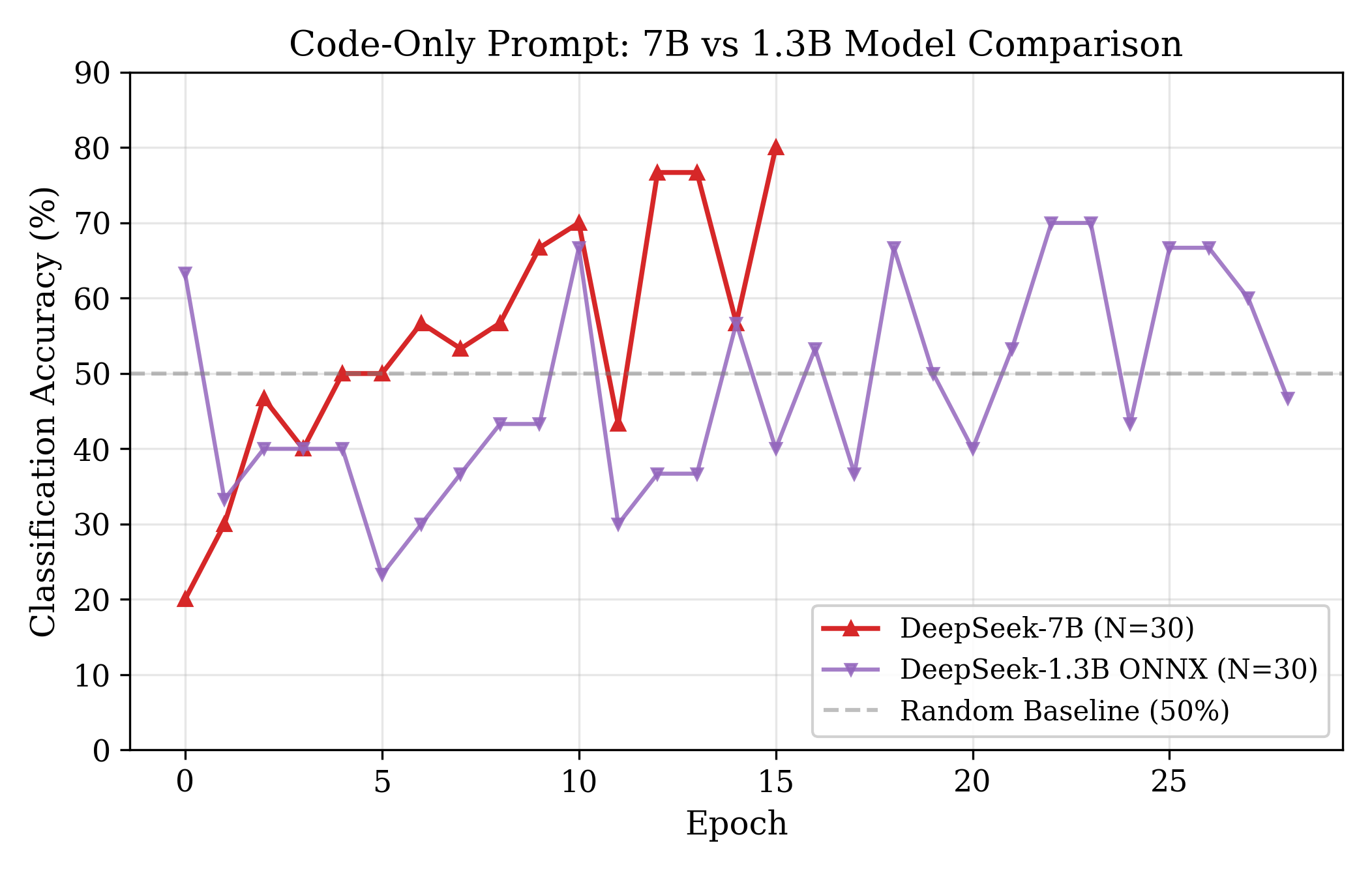}
    \caption{Code-only prompt: DeepSeek-7B (peak 80\%, epoch~15) vs.\ DeepSeek-1.3B with ONNX (peak 70\%, epochs~22--23). The larger model converges faster and more stably.}
    \label{fig:model_comparison}
\end{figure}

\subsubsection{Metadata-Enriched Prompt.}
\begin{figure}[t]
    \centering
    \includegraphics[width=\columnwidth]{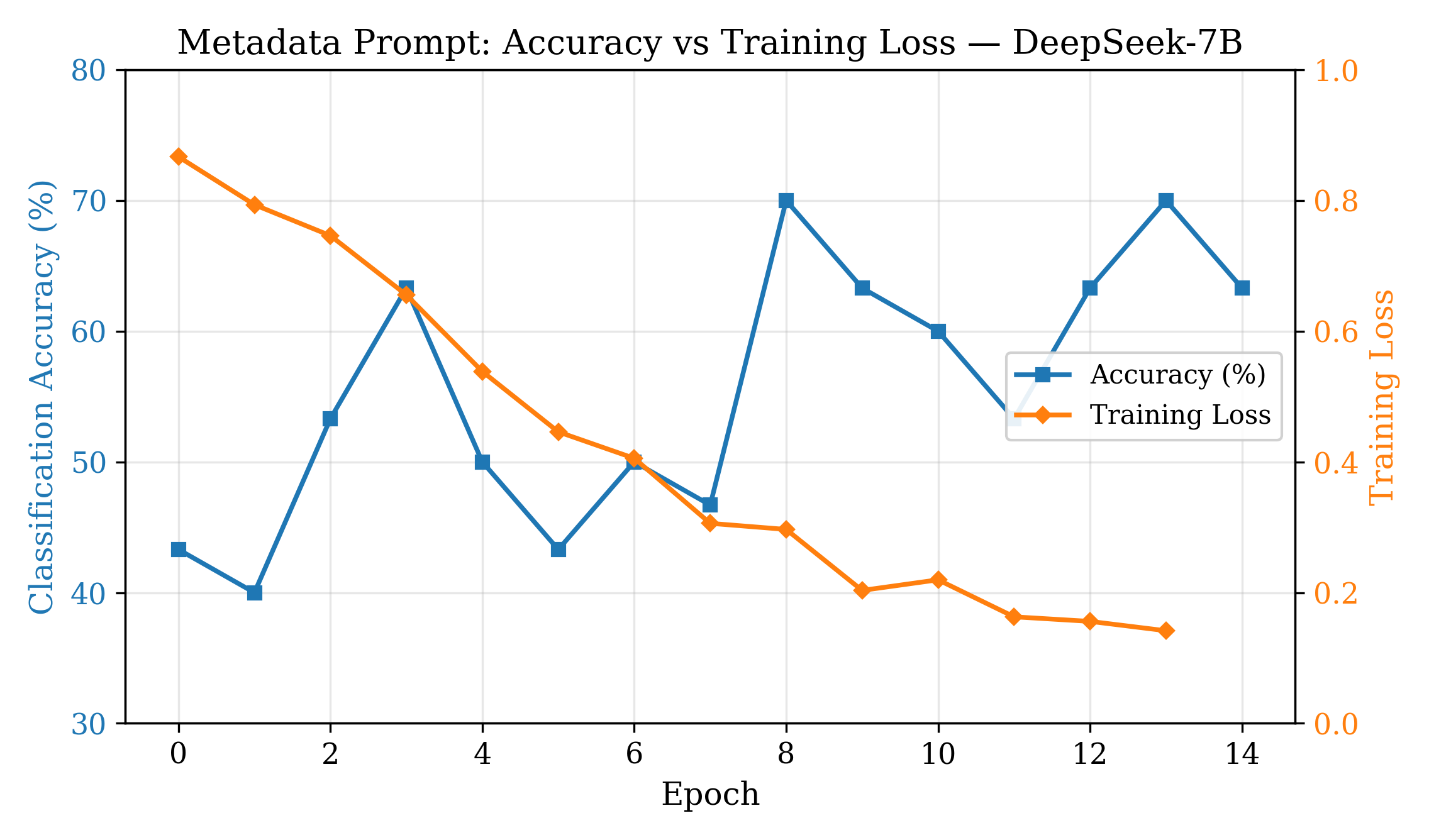}
    \caption{Metadata prompt: accuracy vs.\ training loss. Loss decreases steadily while accuracy oscillates, indicating consistent weight-level learning but a noisy classification signal.}
    \label{fig:meta_acc_loss}
\end{figure}

The metadata prompt achieved 70\% peak accuracy at epochs~8 and~13 with accuracy oscillating between 40\% and 70\%. Training loss decreased steadily from 0.87 to 0.14, indicating consistent weight-level learning, but classification accuracy remained noisy (Fig.~\ref{fig:meta_acc_loss}). This suggests that the mapping from metadata plus code to the correct dataset is inherently more ambiguous than the code-only variant for certain dataset pairs.

\subsection{Per-Dataset Analysis}
\label{sec:per_dataset}

Table~\ref{tab:per_dataset} reveals that the code-only and metadata prompts offer complementary strengths. Metadata excels when dataset properties are distinctive: CelebA-Gender reaches 90.9\% because its metadata (64$\times$64, 2~classes, 162K images) provides an unambiguous fingerprint~\cite{celeba_dataset}. Similarly, SVHN improves to 75.0\% (from 43.1\% code-only) as its metadata is sufficiently unique~\cite{svhn_dataset}. Conversely, CIFAR-100 collapses to 40.5\% and ImageNette to 27.5\% with metadata, both sharing overlapping characteristics with other corpus datasets. The code-only prompt provides more balanced performance across datasets, with CelebA-Gender at 72.5\% and most datasets clustering around 50\%.

\begin{table}[t]
    \centering
    \fontsize{7.5}{9}\selectfont
    \begin{tabular}{lccc}
        \toprule
        \textbf{Dataset} & \makecell{\textbf{Code}\\\textbf{(7B)}} & \makecell{\textbf{Meta}\\\textbf{(7B)}} & \makecell{\textbf{Code}\\\textbf{(1.3B)}} \\
        \midrule
        CelebA-Gender & 72.5 & \textbf{90.9} & 39.2 \\
        CIFAR-10      & 51.9 & 31.2          & \textbf{63.8} \\
        CIFAR-100     & 37.0 & 40.5          & \textbf{49.1} \\
        ImageNette    & 52.4 & 27.5          & \textbf{57.1} \\
        MNIST         & 53.2 & 50.0          & \textbf{53.2} \\
        Places365     & 12.5 & \textbf{45.5} & 10.0 \\
        SVHN          & 43.1 & \textbf{75.0} & 36.1 \\
        \bottomrule
    \end{tabular}
    \caption{Per-dataset aggregated accuracy (\%) by prompt variant and model. Bold indicates the best result per dataset.}
    \label{tab:per_dataset}
\end{table}

\begin{figure}[t]
    \centering
    \includegraphics[width=\columnwidth]{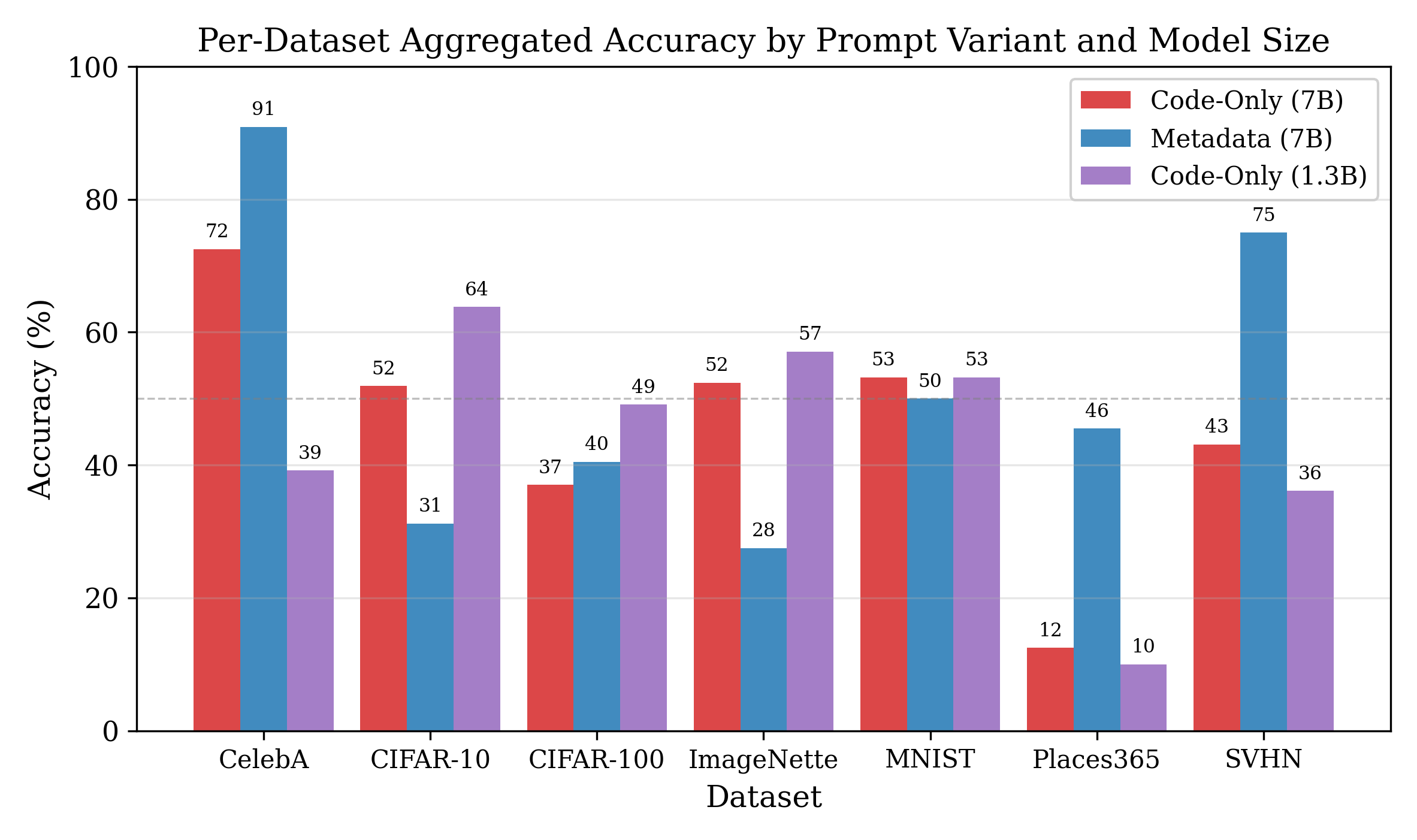}
    \caption{Per-dataset aggregated accuracy across three configurations. Metadata excels for distinctive datasets (CelebA-Gender, SVHN); code-only is more balanced but struggles with Places365.}
    \label{fig:per_dataset}
\end{figure}

\subsection{Ablation Study}
\label{sec:ablation}

The three prompt variants constitute a structured ablation over input information. Removing accuracy values (V1$\to$V2) forces the model to reason from dataset properties rather than numerical comparison, reducing accuracy from 100\% to 70\%. Removing metadata (V2$\to$V3) forces pure code-based reasoning, yet accuracy \emph{increases} to 80\%. This counterintuitive result suggests that metadata may introduce noise for datasets with overlapping properties, while the architecture source code contains richer discriminative signal. The model-size ablation (7B vs.\ 1.3B on code-only) shows that the larger model reaches higher peak accuracy (80\% vs.\ 70\%) with faster convergence (epoch~15 vs.\ 22--23) and substantially lower volatility, confirming that model capacity is a relevant factor for architectural reasoning.

\section{Discussion}
\label{sec:discussion}

The results demonstrate that LLMs can be fine-tuned to predict cross-dataset performance from neural network source code. Both the code-only (80\%) and metadata-enriched (70\%) configurations exceed the 50\% random baseline by a substantial margin, indicating that fine-tuned LLMs extract meaningful discriminative signals.

The finding that code-only outperforms metadata is noteworthy. We hypothesize that architecture source code encodes implicit information about a model's inductive biases---such as receptive field size, depth, and normalization patterns---that correlate with dataset suitability in ways not captured by tabular metadata. Metadata, while informative for distinctive datasets, introduces ambiguity when multiple datasets share similar properties (e.g., CIFAR-10 and SVHN both being 32$\times$32 RGB with 10 classes).

Several limitations should be noted. The evaluation set (N=30) is small; the 10-percentage-point gap between code-only and metadata corresponds to only 3~additional correct predictions. All experiments used a single random seed, so variance across initializations is unknown. The pairwise framing creates statistically dependent samples not controlled for in accuracy computation. The 2,000-character code truncation discards portions of longer architectures, and its effect has not been systematically analyzed. Finally, all experiments are restricted to image classification datasets within LEMUR; generalization to other tasks and datasets remains untested.

\section{Conclusion}
\label{sec:conclusion}

We presented a classification task integrated into the NNGPT framework that extends LLM-based AutoML beyond generative outputs to predicting cross-dataset performance from neural network source code. Using three prompt variants as a structured ablation, we showed that DeepSeek-Coder-7B fine-tuned with LoRA achieves 80\% peak accuracy on the code-only configuration, outperforming the metadata-enriched variant (70\%). Per-dataset analysis revealed complementary strengths: metadata excels for distinctive datasets, while code provides more balanced performance. The model-size comparison (7B vs.\ 1.3B) confirmed that capacity matters for architectural reasoning.

Future work includes evaluating additional LLM families (e.g., CodeLlama, OlympicCoder), held-out generalization splits on unseen architectures and datasets, combining code and metadata in a single prompt, extending to segmentation and detection tasks in LEMUR~\cite{ABrain.NN-Dataset}, and multi-seed runs for statistical robustness.

{
    \small
    \bibliographystyle{ieeenat_fullname}
    \bibliography{bibmain}
}

\end{document}